\documentclass[letterpaper]{article}
\usepackage{aaai2026} 
\usepackage{times} 
\usepackage{helvet} 
\usepackage{courier} 
\usepackage[hyphens]{url} 
\usepackage{graphicx} 
\usepackage{graphicx} 
\usepackage{natbib} 
\usepackage{caption} 
\usepackage{subcaption}
\usepackage{makecell} 
\usepackage{amsfonts}
\usepackage{subcaption}
\usepackage{algorithm}
\usepackage{algorithmic}
\usepackage{booktabs}   
\usepackage{multirow}   
\usepackage{colortbl}   
\usepackage{xcolor}     
\usepackage{adjustbox}
\usepackage{mathrsfs}
\usepackage{amsmath}
\usepackage{amssymb}
\usepackage{amsthm}
\usepackage{mathtools}
\usepackage[mathscr]{eucal}%
\usepackage{newfloat}
\usepackage{listings}
\DeclareCaptionStyle{ruled}{labelfont=normalfont,labelsep=colon,strut=off} 
\floatstyle{ruled}
\newfloat{listing}{tb}{lst}{}
\floatname{listing}{Listing}
\title{HeartLLM: Discretized ECG Tokenization for LLM-Based Diagnostic Reasoning}
\author {
    Jinning Yang\textsuperscript{\rm 1,2},
    Wenjie Sun\textsuperscript{\rm 1,2},
    Wen Shi\textsuperscript{\rm 3, 4}\thanks{Corresponding author}
}
\affiliations {
    \textsuperscript{\rm 1}Shenzhen Institute of Advanced Technology, Chinese Academy of Sciences, Shenzhen, Guangdong, China\\
    \textsuperscript{\rm 2}University of Chinese Academy of Sciences, Beijing, China\\
    \textsuperscript{\rm 3}Department of Neurology, Massachusetts General Hospital, Boston, MA, USA\\
    \textsuperscript{\rm 4}Department of Neurology, Harvard Medical School, Boston, MA, USA\\
    jn.yang@siat.ac.cn, wshi3@mgh.harvard.edu
}

\begin{document}

\maketitle

\begin{abstract}
Electrocardiography (ECG) plays a central role in cardiovascular diagnostics, yet existing automated approaches often struggle to generalize across clinical tasks and offer limited support for open-ended reasoning. We present \textit{HeartLLM}, a novel framework that integrates time-series (TS) and language modeling by enabling large language models (LLMs) to process 12-lead ECG signals for clinical text generation tasks. Our approach discretizes continuous ECG embeddings into quantized codes using a lead-wise encoder and quantization module. These quantized codes are then mapped to an extended ECG vocabulary to form ECG tokens, enabling the model to process both ECG and natural language inputs within a unified framework. To bridge the modality gap, we pretrain the model on an autoregressive ECG token forecasting task, allowing the LLM to capture temporal dynamics through its inherent language modeling capability. Finally, we perform instruction tuning on both ECG question answering and diagnostic report generation. Without modifying the core model, \textit{HeartLLM} achieves strong performance across tasks while maintaining generalization to out-of-distribution settings. Extensive experiments demonstrate the effectiveness of each component and highlight the potential of integrating discretized ECG tokens into LLMs for medical reasoning. 
\end{abstract}

\begin{links}
\link{Code}{https://github.com/yangjinning/HeartLLM}
\end{links}

\section{Introduction}

\begin{figure}[t]
    \centering
    \begin{subfigure}{0.48\linewidth}
        \includegraphics[width=\linewidth]{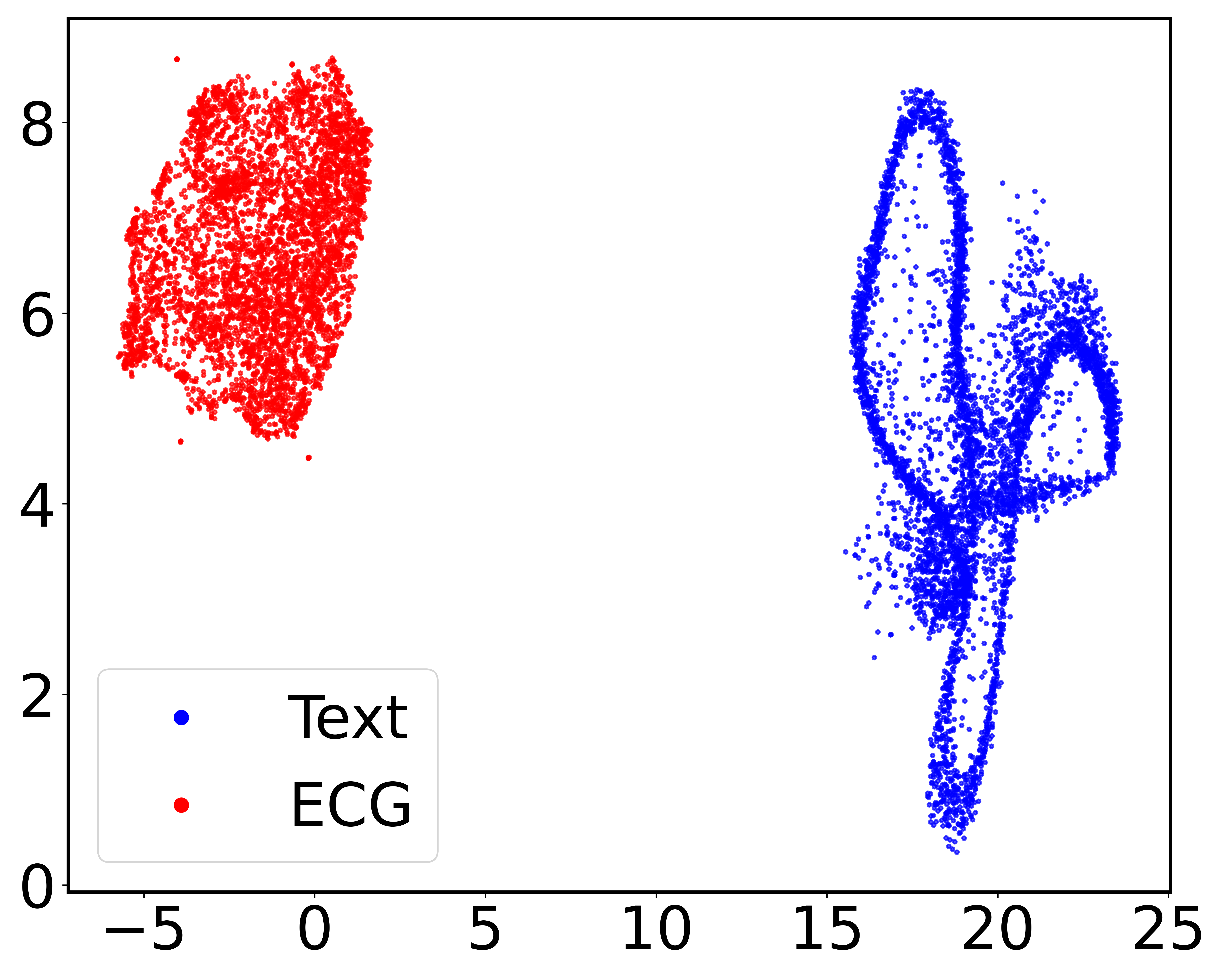}
        \caption{TEST}
    \end{subfigure}
    \hfill
    \begin{subfigure}{0.48\linewidth}
        \includegraphics[width=\linewidth]{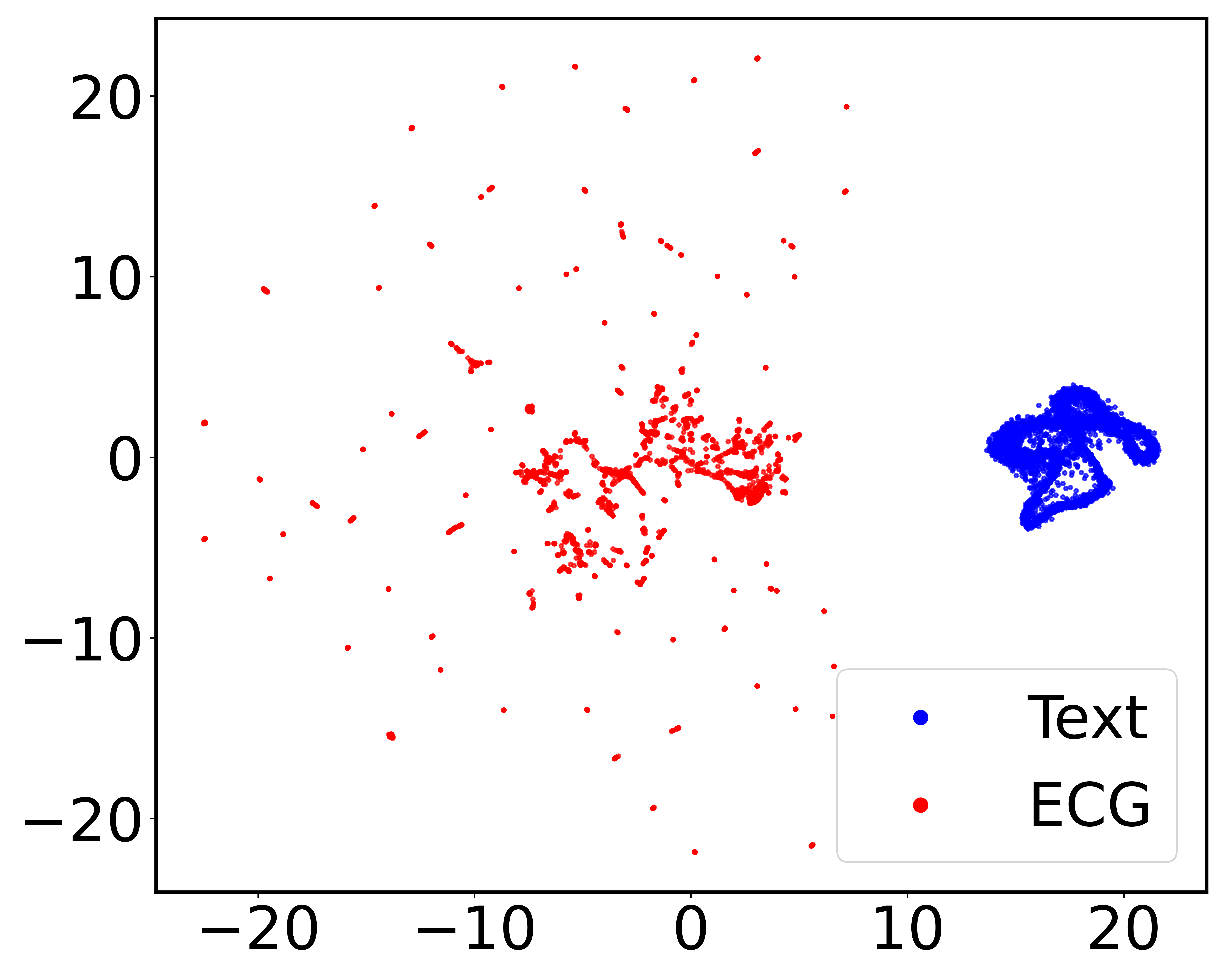}
        \caption{TIMELLM}
    \end{subfigure}

    \begin{subfigure}{0.48\linewidth}
        \includegraphics[width=\linewidth]{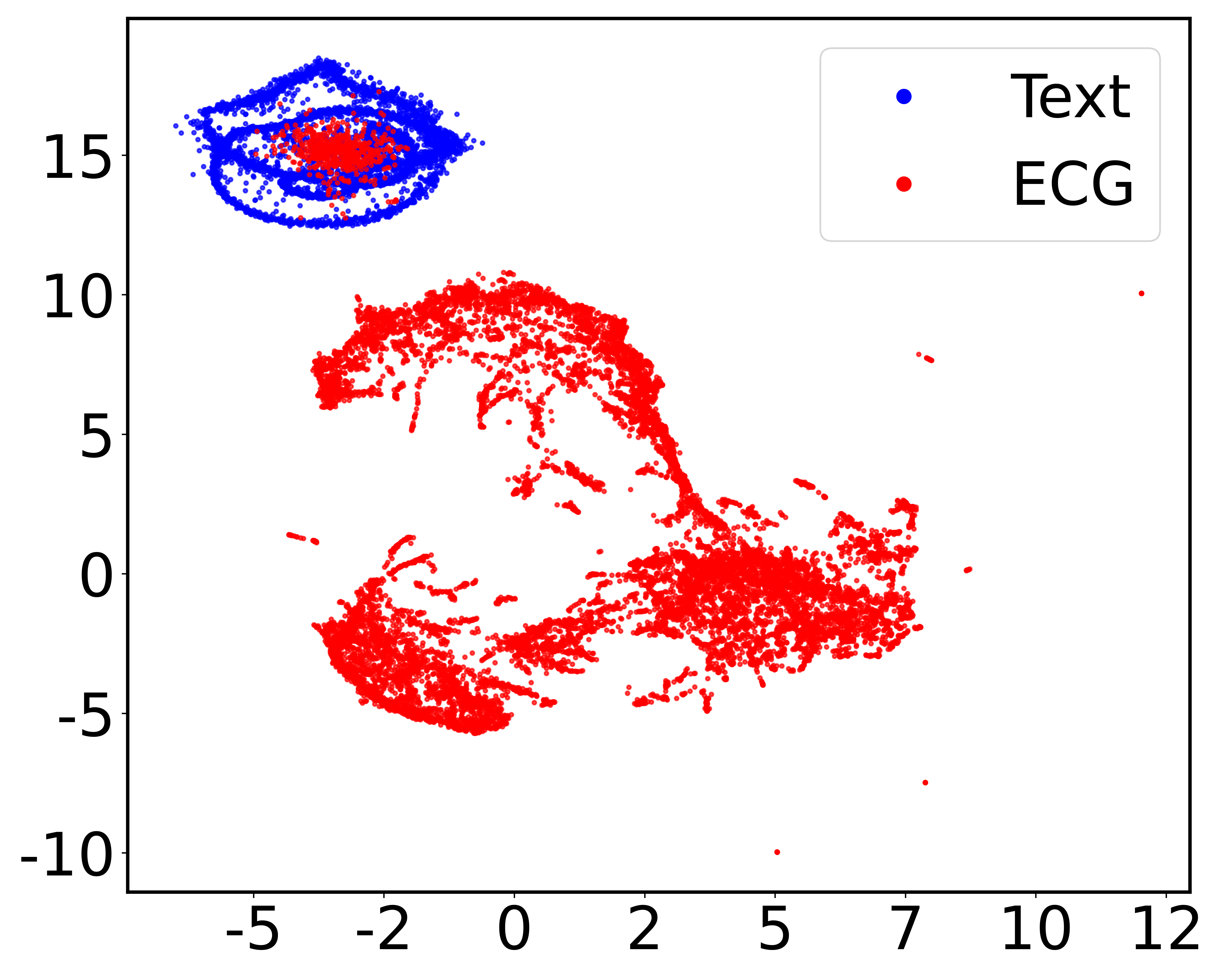}
        \caption{CLIP}
    \end{subfigure}
    \hfill
    \begin{subfigure}{0.47\linewidth}
        \includegraphics[width=\linewidth]{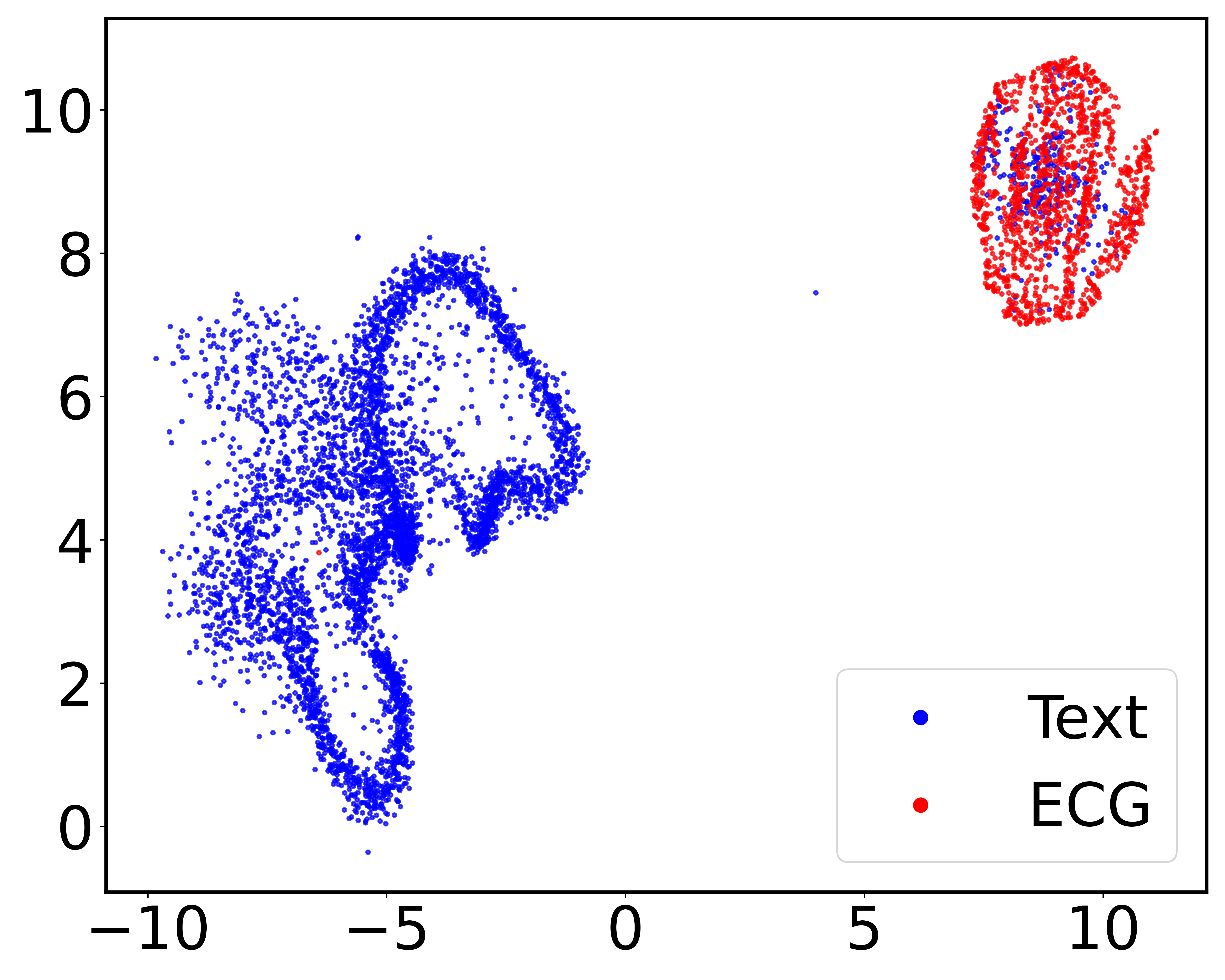}
        \caption{HeartLLM (Ours)}
    \end{subfigure}




    \caption{
    \textbf{UMAP visualizations of ECG and text embedding distributions across different models.} 
    (a-b) TEST and TIMELLM exhibit clear modality separation.
    (c) CLIP shows partially aligned clusters, but still fragmented modality boundaries. 
    (d) HeartLLM aligns all ECG representations with text in a shared semantic space, indicating effective modality unification without explicit contrastive pairing.
    }
    \label{fig:mae_comparison}
\end{figure}

Cardiovascular diseases (CVDs) are the leading cause of mortality and morbidity worldwide~\cite{RN284}, and electrocardiography (ECG) remains a fundamental tool for early diagnosis and monitoring. With the increasing volume and complexity of ECG data in clinical settings, computer-aided interpretation has become an essential supplement to clinical expertise. Recent advances in deep learning have significantly improved performance in clinical decision support and ECG classification tasks~\cite{RN272,xue2023assisting,RN270,RN271,9662723,RN268,xue2023multi}, surpassing traditional signal-processing-based pipelines~\cite{RN281}. However, most existing models are tailored to closed-set classification and rely heavily on supervised training. When confronted with new diagnostic categories or unseen clinical tasks, they require extensive retraining with labeled data, limiting their scalability and adaptability. Moreover, these models are typically designed for narrow classification objectives and cannot support more flexible reasoning tasks such as open-ended question answering or narrative report generation.

In parallel, large language models (LLMs) have demonstrated impressive generalization in vision-language domains~\cite{alayrac2022flamingovisuallanguagemodel,huang2023languageneedaligningperception}, prompting efforts to extend such capabilities to physiological TS data like electrocardiograms (ECGs). However, this extension poses unique challenges: unlike symbolic and semantically structured text, ECG signals are continuous, noisy, and lack explicit semantic boundaries. These fundamental differences complicate the integration of ECG and text into a unified representation space. Recent studies have explored multimodal LLMs for clinical applications, including ECG-based question answering~\cite{liu2024teachmultimodalllmscomprehend,zhao2025ecgchatlargeecglanguagemodel} and report generation~\cite{wan2025meitmultimodalelectrocardiograminstruction}. Most adopt a two-stage paradigm where an ECG encoder produces continuous representations that are subsequently consumed by a language model. However, this design suffers from two major limitations. First, the modality gap between continuous ECG embeddings and discrete textual semantics remains underexplored. Prior alignment strategies in time series–language modeling~\cite{jin2024timellm,sun2024test} or contrastive learning in vision–language settings~\cite{radford2021learningtransferablevisualmodels} often fail to achieve consistent cross-modal alignment. These limitations, when directly applied to ECG, result in either overly clear or fragmented modality boundaries, as illustrated in Figure~\ref{fig:mae_comparison}.
 Second, ECG signals exhibit high redundancy due to waveform repetition and stable low-variance segments. Directly feeding such continuous features into LLMs can lead to overfitting and reduced generalization to rare or subtle signal variations. These challenges highlight the need for a more structured and symbolic interface between ECG signals and language models.

To address the challenges of bridging physiological signals and language models, we introduce \textit{HeartLLM}, a unified framework that enables LLMs to reason over ECG signals in an open-ended, instruction-driven manner. Instead of relying on paired ECG–text supervision or complex cross-modal alignment losses, HeartLLM represents ECGs as discrete symbolic tokens, naturally compatible with the LLM's vocabulary.

Our method is built on three core contributions:
\textit{First}, we design a \textbf{lead-wise encoder} that processes each ECG lead independently using an identical encoder built from temporal inception enhancement blocks. This architecture captures waveform-specific temporal patterns while avoiding inter-lead interference, resulting in precise and interpretable representations.
\textit{Second}, we introduce a \textbf{discretization-based ECG tokenizer} that converts continuous encoder outputs into symbolic tokens via a fixed-scale quantizer (FSQ)~\cite{mentzer2023finite}. These tokens are optimized through autoregressive pretraining and embedded in a shared space with language, enabling seamless integration with LLMs without contrastive learning or paired ECG-text data.
\textit{Third}, we apply \textbf{lightweight instruction tuning} for ECG report generation and question answering. By updating only low-rank modules and prompting with structured clinical inputs, our method achieves strong zero-shot generalization across tasks and datasets with minimal supervision.

HeartLLM is evaluated on two ECG understanding benchmarks: question answering and diagnostic report generation. It achieves state-of-the-art performance across multiple datasets and exhibits strong zero-shot generalization. Further analysis confirms the contribution of each design component, and visualizations illustrate interpretable attention over clinically meaningful waveform regions.

\section{Related Work}

\paragraph{Cross-modal ECG Text Generation}
Recent work has begun to explore the integration of LLMs into ECG interpretation tasks, particularly for open-ended question answering and diagnostic report generation. ECG-ReGen~\cite{tang2024electrocardiogram} adopts a retrieval-augmented framework that identifies similar historical cases to assist report generation in a zero-shot setting. PULSE~\cite{liu2024teachmultimodalllmscomprehend} constructs an instruction-tuning dataset based on ECG images, enabling the fine-tuning of image-text multimodal models. MEIT~\cite{wan2025meitmultimodalelectrocardiograminstruction} introduces TS embeddings directly into the LLM’s attention layers to bridge modality differences, while~\cite{tang2025electrocardiogramlanguagemodelfewshotquestion} treats ECG features as soft prompts and applies meta-learning for rapid adaptation to new tasks. Despite these advances, most existing methods depend on paired ECG–text data to supervise modality alignment, which requires costly annotation and constrains applicability. In contrast, our approach aligns ECG and language representations through token-level pretraining, avoiding the need for paired supervision while maintaining flexibility across tasks.

\paragraph{LLMs for Time Series Tasks}
Large language models have been increasingly applied to TS tasks, typically via two main directions. One leverages LLMs to extract textual features to guide TS modeling~\cite{wang2025ctpdcrossmodaltemporalpattern,liu2025calf,Liu_Xu_Miao_Yang_Zhang_Long_Li_Zhao_2025}. The other treats LLMs as core TS processors, mainly via two paradigms. The TS-as-text approach~\cite{wang2025chattime,ansari2024chronos,wang2024from,NEURIPS2023_3eb7ca52} tokenizes numerical sequences into text for direct input into LLMs, but suffers from length and attention complexity limitations. The TS-embedding-based approach instead encodes signals into continuous embeddings and feeds them into frozen LLMs, often using prompt tuning~\cite{zhou2023one,liu2024autotimes}. Some recent work aligns TS embeddings with language semantics, such as S2IP~\cite{pan2024sipllm}, TEST~\cite{sun2024test}, and TIME-LLM~\cite{jin2024timellm}, through contrastive or reprogramming strategies. Others like VITRO~\cite{inproceedings} and UniTime~\cite{liu2024unitime} propose task-specific vocabularies or reconstruction pretraining. However, most focus on forecasting or classification, with limited support for open-ended language tasks grounded in physiological signals. In contrast, our method discretizes ECG features into symbolic tokens, enabling autoregressive pretraining and unified ECG-text processing for clinical QA and report generation.

\section{Problem Statement}

\begin{figure*}[t]
  \centering
\includegraphics[width=0.95\textwidth]{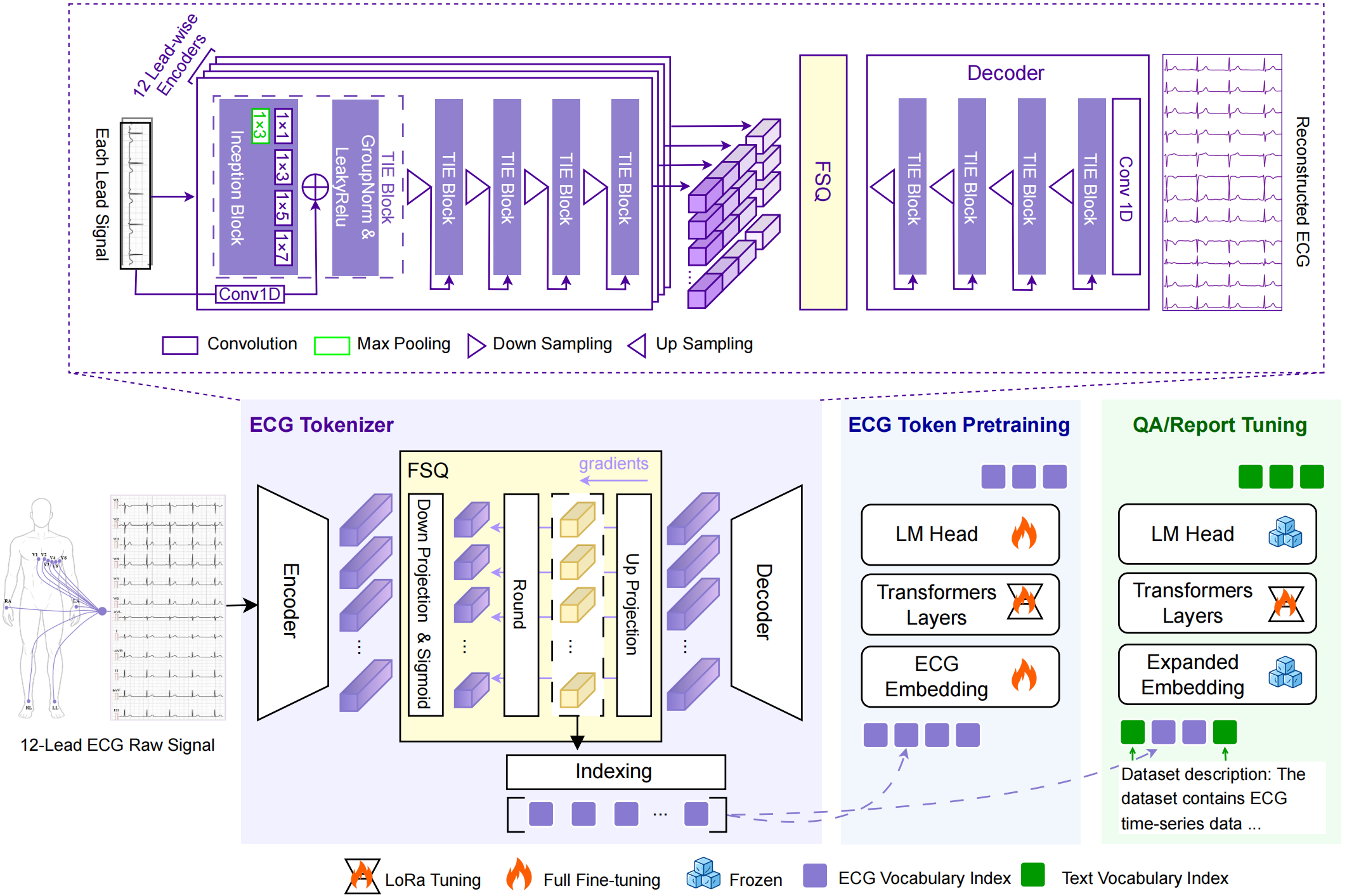}
\caption{Overview of the \textsc{HeartLLM} framework. The model consists of three stages: (1) \textbf{ECG Tokenizer}, where 12-lead ECG signals are encoded by 12 lead-wise encoders and discretized with FSQ into symbolic ECG tokens; (2) \textbf{ECG Token Pretraining}, where an LLM is autoregressively pretrained on ECG tokens to jointly optimize the extended ECG vocabulary and the model using teacher-forcing next-token prediction; and (3) \textbf{QA/Report Tuning}, where the pretrained model is adapted to downstream tasks using lightweight LoRA tuning. Prompts include structured tabular features and textual instructions to guide generation or question answering.}
  \label{fig:framework}
\end{figure*}

Given a collection of 12-lead ECG signals $X = \{ x_i \}_{i=1}^n$, where each $x_i \in \mathbb{R}^{L \times T}$ consists of $L = 12$ leads and $T = 5000$ time steps , we consider two tasks: ECG question answering (ECG-QA) and diagnostic report generation (ECG-Report).

\paragraph{ECG-QA.}
Each signal $x_i$ is paired with a set of natural language questions $Q_i = \{ q_i^{(k)} \}_{k=1}^{m_i}$ and corresponding answers $A_i = \{ a_i^{(k)} \}_{k=1}^{m_i}$. We consider three question types: 
\begin{itemize}
\item \textbf{Verify}: yes/no questions, $A_{\mathrm{verify}} = \{\text{''Yes''}, \text{''No''}\}$.  
Example: ``Does this ECG show any abnormal symptoms? Yes.''
\item \textbf{Choose}: select one or more from predefined options, $A_{\mathrm{choose}} = \{\text{''attr}_1\text{''}, \text{''attr}_2\text{''}, \dots, \text{''none''}\}$.  
Example: ``Which noise does this ECG show, static noise or baseline drift? Baseline drift, static noise.''
\item \textbf{Query}: open-ended, $A_{\mathrm{query}} \subseteq S$, where $S$ is the space of short free-text spans.  
Example: ``What diagnostic symptoms does this ECG show? Incomplete right bundle branch block.''
\end{itemize}
Let $D_{\mathrm{QA}} = \{ (x_i, q_i^{(k)}, a_i^{(k)}) \}_{i,k}$ denote the full set of QA triples. The goal is to learn a model $f_{\mathrm{QA}}: (x_i, q_i^{(k)}) \mapsto \hat{a}_i^{(k)}$ that accurately predicts $\hat{a}_i^{(k)} \in A$.

\paragraph{ECG-Report.}
In this task, the model generates a diagnostic report $r_i$ for each ECG signal $x_i$. The goal is to learn a mapping $f_{\mathrm{report}}: x_i \mapsto \hat{r}_i$ that produces clinically coherent and diagnostically relevant reports.

These two tasks present complementary challenges: ECG-QA requires precise grounding of language in signal features across diverse question formats, while ECG-Report demands the generation of coherent medical narratives from high-dimensional, often noisy physiological input. Successfully addressing both tasks requires the model to align temporal ECG patterns with linguistic representations in a flexible and generalizable manner.

\section{Methods}

\subsection{Overview}

We propose \textit{HeartLLM}, a unified framework that enables large language models to perform open-ended clinical reasoning over ECG signals through symbolic representation and lightweight adaptation.

\textit{First}, we design a lead-wise encoder that captures fine-grained temporal features independently for each ECG lead, enhancing representation fidelity without cross-lead interference. \textit{Second}, we introduce a discretization mechanism that maps continuous features to symbolic tokens using an FSQ, forming an ECG-specific vocabulary compatible with LLMs. These tokens are pretrained via autoregressive prediction, naturally aligning time series and text without paired supervision. \textit{Finally}, HeartLLM is instruction-tuned on clinical tasks such as ECG report generation and question answering, using only LoRA modules for efficient adaptation.
An overview of the full framework is shown in Figure~\ref{fig:framework}.

\subsection{ECG Tokenizer}

Our ECG tokenizer converts a raw 12-leads ECG signal $\mathbf{X} \in \mathbb{R}^{L \times T}$ into a sequence of discrete tokens suitable for language modeling. This process includes three components: (1) 12 lead-wise encoders that extract multi-scale TS representations from each lead, which are then aggregated to capture the global cross‑lead context; (2) a fixed-scale quantizer (FSQ) that first compresses TS representations into a low-dimensional latent space, discretizes this latent space to obtain quantized codes, and finally dequantizes the codes back into the original TS representation latent space; (3) a decoder that mirrors the encoder architecture and, with a self‑supervised reconstruction loss, ensures that the learned tokens preserve physiological information. The quantized codes are first converted into scalar indices, which are subsequently mapped to the ECG vocabulary. This vocabulary extends the original vocabulary of the LLM, allowing for unified modeling of text and ECG signals.

\paragraph{Lead-wise Encoder.}  
Each lead $x_\ell$ is first z-normalized and independently encoded using a stack of convolutional blocks. The encoder consists of two key components: an {Inception Block} to capture diverse temporal patterns and a {Temporal Inception Enhancement (TIE) Block} for deep contextual modeling.

\textit{Inception Block.}  
The Inception Block applies multiple 1D convolutions with kernel sizes $\{1, 3, 5, 7\}$ in parallel to extract multi-scale temporal features. 

\textit{Temporal Inception Enhancement (TIE) Block.}  
The outputs from the Inception Block are summed and passed through GroupNorm and a LeakyReLU activation, followed by a residual connection to preserve information flow:
\[
\mathbf{h}^{(i)} = \mathrm{LeakyReLU}\left(\mathrm{GN}\left(\sum_{k} \mathrm{Conv1D}_k(\mathbf{h}^{(i-1)})\right) + \mathbf{h}^{(i-1)}\right).
\]
TIE blocks are applied sequentially to capture long-range signal dependencies, followed by convolutional downsampling layers that progressively compress the temporal resolution. This hierarchical structure allows the encoder to balance fine-grained waveform preservation with contextual abstraction. The encoded representations from all lead-wise encoders are concatenated along the lead dimension to form the final latent ECG representation $\mathbf{Z} \in \mathbb{R}^{T' \times D'}$.

\paragraph{Fixed-Scale Quantization (FSQ).} 
To discretize continuous ECG representations, each latent ECG representation at timestep $t$  $\mathbf{z}_t \in \mathbb{R}^{D'}$ is projected into a lower-dimensional space $\mathbb{R}^{D}$ using a linear layer with parameters $\mathbf{W} \in \mathbb{R}^{D \times D'}$ and $\mathbf{b} \in \mathbb{R}^{D}$:
\[
\mathbf{z}_t^{\mathrm{cont}} = \sigma(\mathbf{W} \mathbf{z}_t + \mathbf{b}),
\]
where $\sigma(\cdot)$ is the sigmoid function. The quantized code is obtained by uniformly quantizing each projected dimension into $K$ discrete levels:
\[
\mathbf{z}_t^{\mathrm{disc}} = \frac{1}{K - 1} \cdot \mathrm{round} \left( \mathbf{z}_t^{\mathrm{cont}} \cdot (K - 1) \right),
\]
with rounding applied element-wise. A straight-through estimator~\cite{bengio2013estimatingpropagatinggradientsstochastic} is used to enable gradient flow through the quantization step. The quantized code $\mathbf{z}_t^{\mathrm{disc}}$ is then projected back to the latent space of $\mathbf{z}_t$, yielding $\mathbf{z}'_t \in \mathbb{R}^{D'}$.

\paragraph{Token Indexing and Vocabulary Construction.} 
To generate symbolic ECG tokens, the quantized code $\mathbf{z}_t^{\mathrm{disc}}$ is mapped to a unique token index using positional base conversion:
\[
\mathit{token\_id}_t = \sum_{i=1}^{D} v_{t,i} \cdot K^{i-1},
\]
where $v_{t,i} = \mathrm{round}\left( z_{t,i}^{\mathrm{disc}} \cdot (K -1) \right) \in \{0, 1, \dots, K -1\}$.
This yields $\mathit{token\_id}_t \in \{0, \dots, K^D - 1\}$. These indices are used to look up embedding vectors from an ECG-specific vocabulary:
\[
\mathrm{tok}^{ECG}_t = \mathrm{ECG\_Vocabulary}[\mathit{token\_id}_t],
\]
where $\mathrm{ECG\_Vocabulary} \in \mathbb{R}^{|\mathcal{V}_{\mathrm{ECG}}| \times d_{LLM}}$ and $|\mathcal{V}_{\mathrm{ECG}}| = K^D$.

\paragraph{Decoder and Autoencoder Objective.} 
To ensure that discretization preserves physiologically meaningful information, we employ a symmetric decoder mirroring the encoder structure. It uses transposed 1D convolutions for temporal upsampling and stacked TIE blocks to reconstruct fine-grained waveform details. The reconstruction is trained using a mean squared error loss:
\[
\hat{X} = f_{\mathrm{dec}}(\mathbf{Z'}), \quad
\mathcal{L}_{\mathrm{recon}} = \frac{1}{LT} \sum_{\ell=1}^{L} \sum_{t=1}^{T} \left( x_{\ell,t} - \hat{x}_{\ell,t} \right)^2.
\]

\paragraph{ECG Tokenization for Language Modeling.} 
The resulting ECG token sequence $\{ \mathrm{tok}^{ECG}_t \}_{t=1}^{T'}$ can be used as input to a language model. ECG tokens are initialized from the same distribution as the LLM’s text tokens. In the next stage, we leverage these ECG token sequences for autoregressive pretraining, using the same pretraining objective as for text tokens. Therefore, ECG token embeddings are drawn from the same embedding space as text tokens, the LLM can process ECG and text in a unified fashion.

\subsection{Autoregressive Pretraining on ECG Tokens} 
To integrate ECG signals into the language modeling framework, we pretrain the model on an autoregressive forecasting task over discretized ECG tokens. This allows the LLM to build temporal dependencies within ECG sequences and learn to embed symbolic ECG tokens into its semantic space.

Given an input ECG signal $\mathbf{X}$, the tokenizer converts it into a discrete token sequence $\{ \mathrm{tok}^{ECG}_t \}_{t=1}^{T'}$, where each token is embedded in the same space as textual input. During pretraining, a random slice of this sequence is selected and split into a historical context and a prediction target (typically in a 9:1 ratio). The LLM is trained to autoregressively predict the next ECG tokens from the context:

\[
\mathcal{L}_{\mathrm{AR}}
= \frac{1}{\sum_{i=1}^{N} |\mathcal{M}_i|}
\sum_{i=1}^{N}\sum_{t \in \mathcal{M}_i}
\mathrm{CE}\!\left(
  \mathrm{LM}_\theta(\mathrm{tok}_{i,\le t}^{ECG}),
  \mathrm{tok}_{i,t+1}^{ECG}
\right),
\label{eq:ar_loss}
\]
where $N$ is the batch size, $\mathcal{M}_i$ is the set of prediction steps, $\mathrm{LM}_\theta(\cdot)$ is the LLM output logits over the ECG vocabulary, and $\mathrm{CE}(\cdot,\cdot)$ is the cross-entropy loss.

\paragraph{Training Strategy.}
The pretraining data is constructed from the MIMIC-IV-ECG database~\cite{gow2023mimic}, yielding over 1.5 million time-series slices. To enable parameter-efficient adaptation, we fine-tune only a subset of the LLM parameters: the ECG embedding table and output classification head are fully updated, while the linear projections within the transformer blocks (e.g., QKV and feedforward layers) are adapted using low-rank LoRA modules. This strategy allows the model to efficiently acquire ECG-specific temporal structure while preserving the general language capabilities of the backbone model.

\subsection{Instruction Tuning for Clinical Tasks} 
In the final stage, we fine-tune the model on downstream clinical tasks, including ECG diagnostic report generation and open-ended question answering. This stage enables the pretrained model from the previous stage to align symbolic ECG representations with textual outputs in task-specific contexts. Following common instruction-tuning practice, the embedding table and output classification head are frozen, and only the linear projection layers of the LLM (e.g., QKV and feed-forward layers) are updated with LoRA.

Each training sample is structured as a prompt that concatenates five components: (1) a brief dataset description (e.g., “The dataset contains electrocardiogram (ECG) time-series data.”), (2) a task-specific instruction (e.g., “Based on the given ECG signal embeddings, generate an ECG diagnostic report.”), (3) tabular features (including patient ID, age, sex, height, weight, recording date, nurse ID, device ID, and recording site),  (4) an optional natural language question when the task is ECG-QA, and (5) ECG indicators (e.g., \texttt{<|start\_ecg|>} and \texttt{<|end\_ecg|>}) that mark the position for inserting ECG token sequences. The model is trained to autoregressively generate either a diagnostic report or an answer span, depending on the task. The supervision signal comes from expert-written reports or curated answers, enabling the model to learn both structured reasoning and domain-specific terminology. This unified format allows the model to flexibly handle both classification-type questions and free-form report generation, without needing task-specific heads or architectural changes.

\begin{table*}[t]          
  \setlength{\tabcolsep}{5pt}
  \centering
\scalebox{0.8}{
    \begin{tabular}{c c c|*{10}{c}} 
    \toprule
    \multirow[c]{2}{*}{\textbf{Dataset}} &
    \multirow[c]{2}{*}{\textbf{Task}}    &
    \multirow[c]{2}{*}{\textbf{Metric}}  &
    \multicolumn{10}{c}{\textbf{Models}} \\
    \cmidrule(lr){4-13}
    & & &
    \textbf{Ours} &
    TIMELLM &
    TEST &
    MEIT &
    STMEM &
    \makecell{W2V-\\CMSC-\\RLM} &
    LLMTIME &
    ChatTime &
    \makecell{Fusion\\Trans.} &
    M3AE \\
    \midrule
    \multirow{4}{*}{\rotatebox[origin=c]{90}{\textbf{\shortstack{MIMIC-\\IV-ECG}}}} &
      QA-Verify  & EM~ACC & \textbf{78.12} & 66.00 & 67.86 & 67.69 & 69.73 & 71.57 & 69.97 & 68.78 & \underline{72.91} & -- \\
    & QA-Choose  & EM~ACC & \textbf{69.73} & 28.86 & 41.21 & 50.18 & 49.39 & \underline{65.80} & 55.77 & 52.37 & 64.38 & -- \\
    & QA-Query   & EM~ACC & \textbf{22.05} &  3.24 &  8.65 &  6.67 &  8.22 & \underline{15.24} &  8.16 &  6.26 & 14.35 & -- \\
    & QA-Average & EM~ACC & \textbf{56.63} & 32.70 & 39.24 & 41.51 & 42.45 & \underline{50.87} & 44.63 & 42.47 & 50.55 & -- \\
    \midrule
    \multirow{4}{*}{\rotatebox[origin=c]{90}{\textbf{PTB-XL}}}&
      QA-Verify  & EM~ACC & \textbf{72.66} & 67.46 & 66.43 & 65.34 & 68.70 & \underline{71.48} & 67.96 & 67.65 & 64.27 & 67.73 \\
    & QA-Choose  & EM~ACC & \textbf{59.74} & 29.42 & 47.25 & 48.06 & 46.61 & \underline{58.66} & 45.89 & 43.19 & 50.95 & 31.15 \\
    & QA-Query   & EM~ACC & \textbf{32.55} &  9.24 & 23.58 & 14.89 & 22.62 & \underline{31.57} & 20.15 & 18.98 & 25.30 & 22.71 \\
    & QA-Average & EM~ACC & \textbf{54.98} & 35.37 & 45.75 & 42.76 & 45.98 & \underline{53.90} & 44.67 & 43.27 & 46.84 & 40.53 \\
    \midrule
    \multirow{8}{*}{\rotatebox[origin=c]{90}{\textbf{MIMIC-IV-ECG}}} &
    \multirow{8}{*}{Report} & BLEU-1   & \textbf{44.99} & 14.49 & 28.40 & 27.20 & 24.89 & \underline{35.89} & 28.92 & 26.29 & -- & -- \\
    & & BLEU-2  & \textbf{37.61} & 7.68 & 19.50 & 18.78 & 15.92 & \underline{28.00} & 19.71 & 16.92 & -- & -- \\
    & & BLEU-3  & \textbf{33.59} & 5.69 & 15.51 & 14.73 & 12.73 & \underline{24.14} & 15.84 & 12.98 & -- & -- \\
    & & BLEU-4  & \textbf{29.74} & 3.52 & 12.24 & 11.23 &  9.83 & \underline{20.60} & 12.27 &  9.62 & -- & -- \\
    & & METEOR  & \textbf{46.99} & 15.20 & 28.76 & 31.45 & 25.26 & \underline{37.33} & 29.15 & 26.86 & -- & -- \\
    & & ROUGE-1 & \textbf{58.67} & 26.57 & 41.85 & 39.42 & 36.43 & \underline{49.06} & 41.81 & 37.96 & -- & -- \\
    & & ROUGE-2 & \textbf{45.41} & 11.86 & 25.68 & 23.72 & 20.45 & \underline{35.02} & 25.50 & 21.79 & -- & -- \\
    & & ROUGE-L & \textbf{58.63} & 26.39 & 41.82 & 39.41 & 36.40 & \underline{49.04} & 41.77 & 37.92 & -- & -- \\
    \midrule
    \multirow{8}{*}{\rotatebox[origin=c]{90}{\textbf{PTB-XL}}} &
    \multirow{8}{*}{Report} & BLEU-1   & \textbf{49.61} & 21.03 & 32.37 & 17.82 & 38.07 & \underline{47.65} & 22.18 & 21.38 & -- & -- \\
    & & BLEU-2  & \textbf{43.35} & 13.02 & 25.42 & 12.27 & 31.86 & \underline{40.98} & 16.12 & 15.75 & -- & -- \\
    & & BLEU-3  & \textbf{37.50} & 5.75 & 20.74 &  8.54 & 25.86 & \underline{34.87} & 13.28 & 12.98 & -- & -- \\
    & & BLEU-4  & \textbf{33.60} & 4.22 & 17.83 &  6.17 & 22.20 & \underline{30.92} & 12.03 & 11.82 & -- & -- \\
    & & METEOR  & \textbf{53.90} & 23.94 & 34.14 & 25.14 & 41.11 & \underline{53.33} & 23.64 & 23.12 & -- & -- \\
    & & ROUGE-1 & \textbf{59.53} & 38.74 & 47.27 & 34.94 & 49.39 & \underline{57.08} & 38.35 & 37.58 & -- & -- \\
    & & ROUGE-2 & \textbf{45.12} & 18.24 & 31.68 & 19.58 & 34.20 & \underline{42.31} & 23.37 & 22.86 & -- & -- \\
    & & ROUGE-L & \textbf{59.24} & 38.69 & 47.01 & 34.81 & 49.30 & \underline{56.72} & 38.29 & 37.56 & -- & -- \\
    \bottomrule
  \end{tabular}
  }
  
\caption{Evaluation results on ECG question answering and diagnostic report generation tasks. All metrics are reported as percentages (\%), where higher values indicate better performance. For each dataset, the best-performing result in each task is highlighted in \textbf{bold}, while the second-best is \underline{underlined}.}
  \label{tab:main_results}
\end{table*}

\section{Experiments}
\subsection{Experimental Settings}
\paragraph{Datasets and Tasks.}
We evaluate our method on two ECG-related language generation tasks: ECG-based question answering (ECG-QA) and diagnostic report generation (ECG-Report). The ECG-QA task follows the setup of~\cite{tang2024electrocardiogram}, where 10,598 questions are curated based on six types of ECG attributes—extra systole, heart axis, signal noise, numerical indicators, SCP codes, and infarction stages—using recordings from PTB-XL~\cite{RN290} and MIMIC-IV-ECG~\cite{gow2023mimic}. Only single-ECG questions are considered. The ECG-Report task uses the same datasets with official diagnostic reports. Each dataset is split into training, validation, and test sets in a 7:1:2 ratio by patient ID to prevent data leakage. For MIMIC-IV-ECG, only ECGs without missing values are retained, and one ECG is randomly selected per patient, resulting in 78,358 ECGs. PTB-XL contains 21,797 ECGs.

\paragraph{Baselines.}
We compare our method with a range of TS-Text multimodal models. These include two TS-as-text baselines: ChatTime~\cite{wang2025chattime} and LLMTIME~\cite{NEURIPS2023_3eb7ca52}, which directly feed raw numerical sequences into language models. We further consider five TS-embedding-based methods: TIME-LLM~\cite{jin2024timellm}, TEST~\cite{sun2024test}, MEIT~\cite{wan2025meitmultimodalelectrocardiograminstruction}, and LLaVA~\cite{liu2023visualinstructiontuning} with either ST-MEM~\cite{na2024guiding} or W2V-SMSC-RLM~\cite{RN268} as the ECG encoder. In addition, we include two strong multimodal baselines specific to ECG-QA: Fusion Transformer~\cite{oh2023ecgqa} and M3AE~\cite{chen2022multi}. Note that Fusion Transformer and M3AE do not support text generation and are thus excluded from ECG-Report evaluation. Moreover, M3AE requires pretraining with classification labels, which are not available in MIMIC-IV-ECG.

\paragraph{Evaluation Metrics.}
We use exact match accuracy (EM ACC) to evaluate ECG-QA~\cite{tang2024electrocardiogram}. For ECG-Report, standard text generation metrics are reported, including BLEU~\cite{papineni-etal-2002-bleu}, METEOR~\cite{banerjee-lavie-2005-meteor}, and ROUGE-F1~\cite{lin-2004-rouge}.

\paragraph{Implementation Details.}
All models based on large language models use LLaMA-3.2-3B as the default backbone, except ChatTime, which is built on LLaMA-2-7B. For methods involving LLM tuning, we apply 4-bit quantization with LoRA. LoRA rank is set to 64 with a scaling factor of 16. Models are trained for one epoch with a constant learning rate of 1e-4 and a batch size of 10. For ECG discretization, the number of discrete levels is set to \(K = 6\) and the dimension of the quantized code to \(D = 4\), resulting in an ECG vocabulary of 1296 unique tokens. All experiments are conducted on one A100 GPU.

\subsection{Main Results} 

\paragraph{Comparison with SOTA}
Table~\ref{tab:main_results} summarizes the performance of HeartLLM and various baselines across ECG-QA and ECG-Report tasks on the MIMIC-IV-ECG and PTB-XL datasets. 
For ECG-QA, our model consistently achieves the highest exact match accuracy on both datasets, reaching \textbf{56.63} on MIMIC-IV-ECG and \textbf{54.98} on PTB-XL, with significant margins especially in the \textit{Query} setting, which poses the greatest semantic challenge. 
This indicates that HeartLLM can more effectively capture and interpret the clinical semantics of ECG signals compared to both TS-as-text and embedding-based LLM methods. 
In the ECG-Report task, HeartLLM also obtains the best results across all evaluation metrics.
On PTB-XL, it achieves a BLEU-4 score of \textbf{33.60}, METEOR of \textbf{53.90}, and ROUGE-L of \textbf{59.24}, outperforming competitive models like TIME-LLM and LLaVA+W2V-CMSC-RLM.
Our method also achieves strong performance on MIMIC-IV-ECG in a zero-shot setting, indicating robust generalization across datasets. These results validate the advantage of our discretization-based representation and token-level alignment between TS and language modalities.

\subsection{Ablation Study}
\begin{figure}[t]
  \centering
  \begin{tabular}{cc}
    \begin{subfigure}{0.48\columnwidth}
      \includegraphics[width=\linewidth]{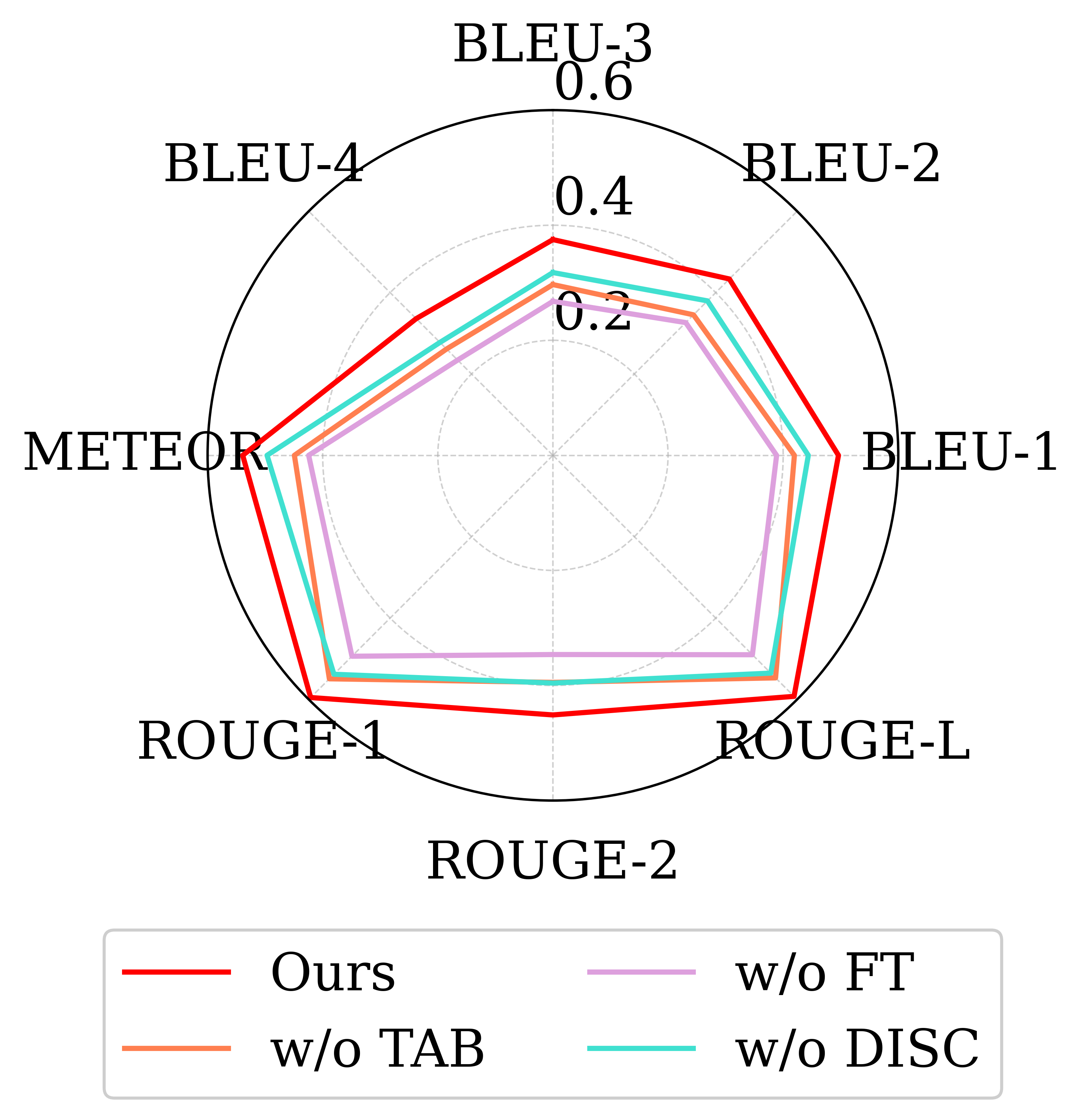}
      \caption{PTB-XL Report}
    \end{subfigure} &
    \begin{subfigure}{0.40\columnwidth}
      \includegraphics[width=\linewidth]{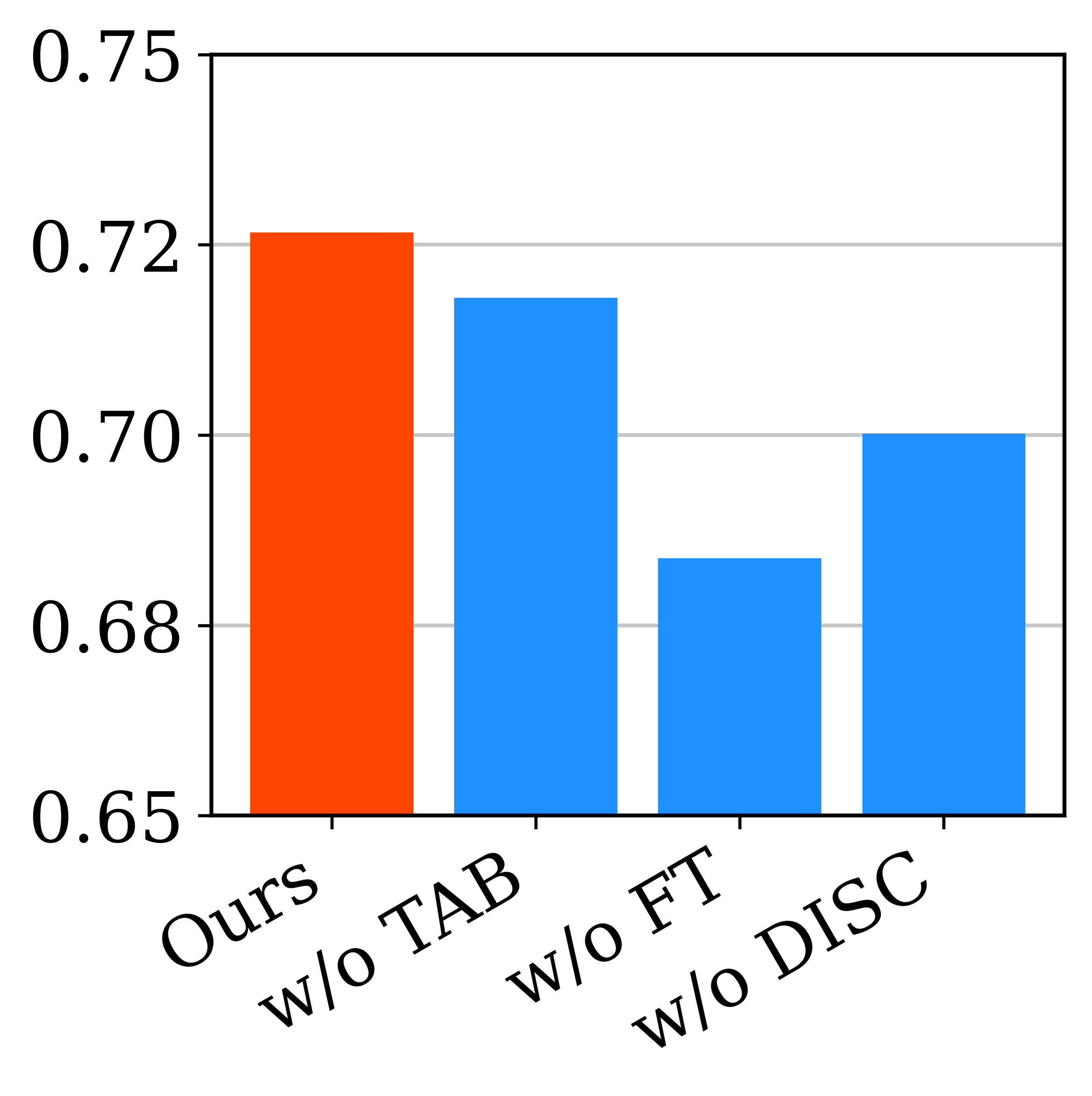}
      \caption{PTB-XL QA-Verify}
    \end{subfigure} \\
    \begin{subfigure}{0.40\columnwidth}
      \includegraphics[width=\linewidth]{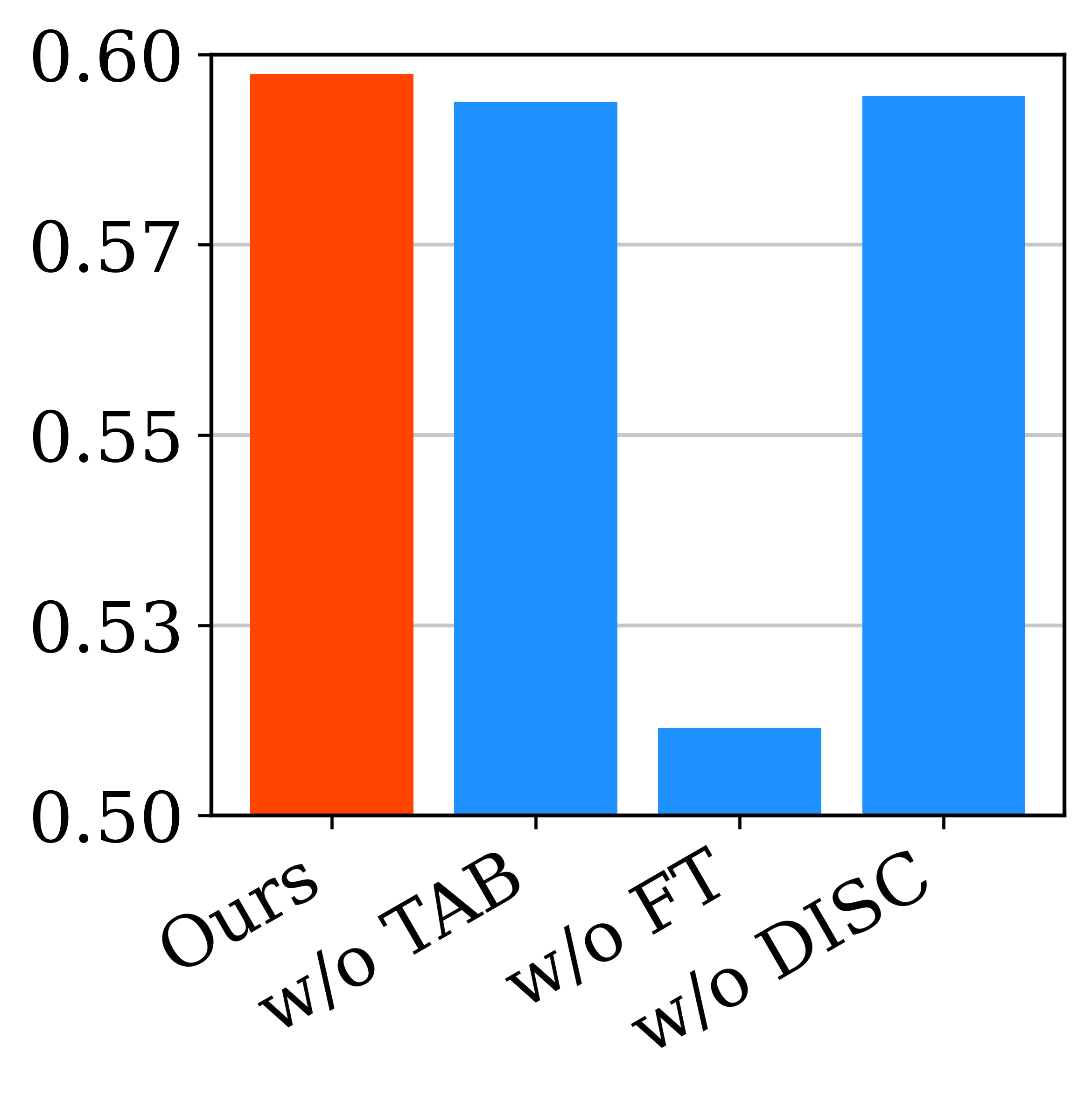}
      \caption{PTB-XL QA-Choose}
    \end{subfigure} &
    \begin{subfigure}{0.40\columnwidth}
      \includegraphics[width=\linewidth]{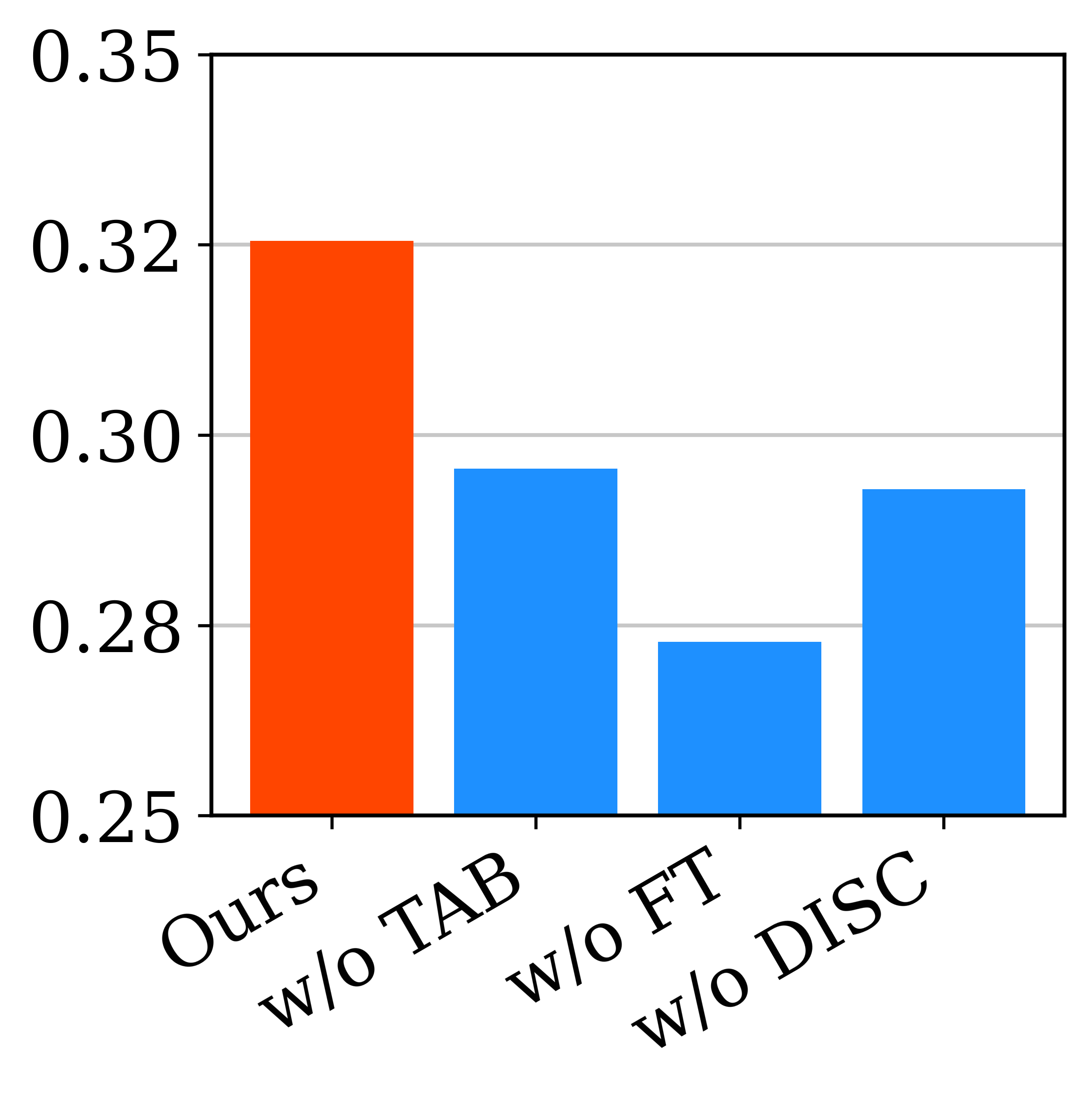}
      \caption{PTB-XL QA-Query}
    \end{subfigure}
  \end{tabular}
  \caption{Ablation study on the PTB-XL. (a) Report generation evaluated by BLEU, METEOR, and ROUGE. (b–d) EM accuracy on QA-Verify, QA-Choose, and QA-Query tasks. We compare the full model (Ours) against three variants: w/o TAB, w/o FT, and w/o DISC.}
  \label{fig:ab_results}
\end{figure}

We conduct ablations on PTB-XL to isolate the effects of three core components in \textsc{HeartLLM}. Results are shown in Figure~\ref{fig:ab_results}. Removing the discretization module (w/o DISC) causes consistent drops across all tasks, suggesting that symbolic ECG tokens are more compatible with LLMs than continuous embeddings. Without fine-tuning the LLM (w/o FT), performance notably declines—especially on QA-Choose—indicating that LoRA-based adaptation helps the model capture task-specific semantics. Excluding tabular features in the prompt (w/o TAB) slightly weakens generation and QA-Verify, reflecting their complementary role in providing patient context. Together, these results confirm that each component contributes meaningfully to downstream performance.

\subsection{Analysis}
\paragraph{Effectiveness of Our Encoder.} 
We compare our lead-independent ResIncept encoder with four representative alternatives: MultiCNN~\cite{oh2023ecgqa}, used in Fusion Transformer and MEIT; CausalCNN~\cite{sun2024test}, the default encoder in TEST; Unet2D, a 2D convolution-based U-Net to extract channel-dependent features; and IndUnet1D, a structure that applies independent 1D U-Nets to each lead. As shown in Table~\ref{tab:ecg_encoder_comparison}, our encoder consistently outperforms all baselines across both datasets and all QA tasks. Notably, it yields the largest gains on QA-Query, which requires nuanced reasoning over subtle waveform features. Compared to MultiCNN and Unet2D, which either couple all leads early, our encoder better preserves fine-grained temporal features while maintaining lead-specific context. The improvement over CausalCNN suggests that fixed dilation patterns are less effective than our multi-scale design for ECGs.

\begin{table}[t]
\centering
\setlength{\tabcolsep}{3pt}

{\fontsize{7}{8}\selectfont
\begin{tabular}{ccccccc}
\toprule
\textbf{Dataset} & \textbf{Task} & \textbf{Unet2D} & \textbf{Unet1D} & \textbf{MultiCNN} & \textbf{CausalCNN} & \textbf{Ours} \\
\midrule
\multirow{3}{*}{\makecell[c]{\textbf{MIMIC-}\\\textbf{IV-ECG}}}
& QA-Verify  & 75.09 & 76.58 & 73.09 & 75.20 & \textbf{78.12} \\
& QA-Choose  & 67.29 & 68.01 & 63.66 & 68.07 & \textbf{69.73} \\
& QA-Query   & 18.71 & 20.28 & 16.71 & 18.74 & \textbf{22.05} \\
\midrule
\multirow{3}{*}{\textbf{PTB-XL}} 
& QA-Verify  & 71.47 & 70.89 & 69.53 & 70.54 & \textbf{72.66} \\
& QA-Choose  & 59.49 & 57.02 & 54.38 & 59.08 & \textbf{59.74} \\
& QA-Query   & 29.49 & 31.14 & 29.50 & 29.45 & \textbf{32.55} \\
\bottomrule
\end{tabular}
}

\caption{Performance comparison of different ECG encoders on QA tasks across MIMIC-IV-ECG and PTB-XL datasets. Each score indicates the EM ACC (\%).}
\label{tab:ecg_encoder_comparison}
\end{table}

\begin{figure}[t]
  \centering
  \begin{subfigure}[t]{0.30\linewidth}
    \centering
    \includegraphics[height=4.9cm]{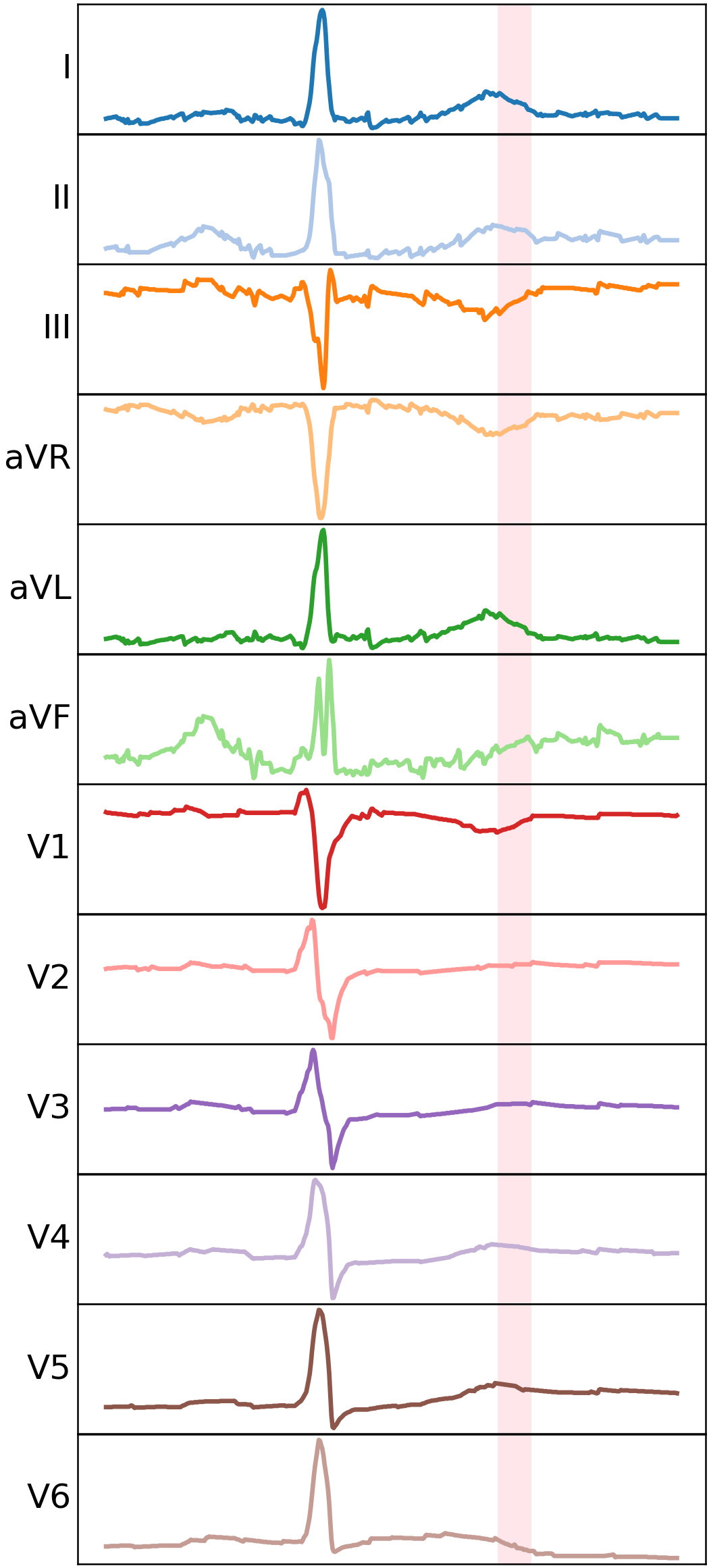}
    \caption{}
  \end{subfigure}
  \hfill
  \begin{subfigure}[t]{0.30\linewidth}
    \centering
    \includegraphics[height=4.9cm]{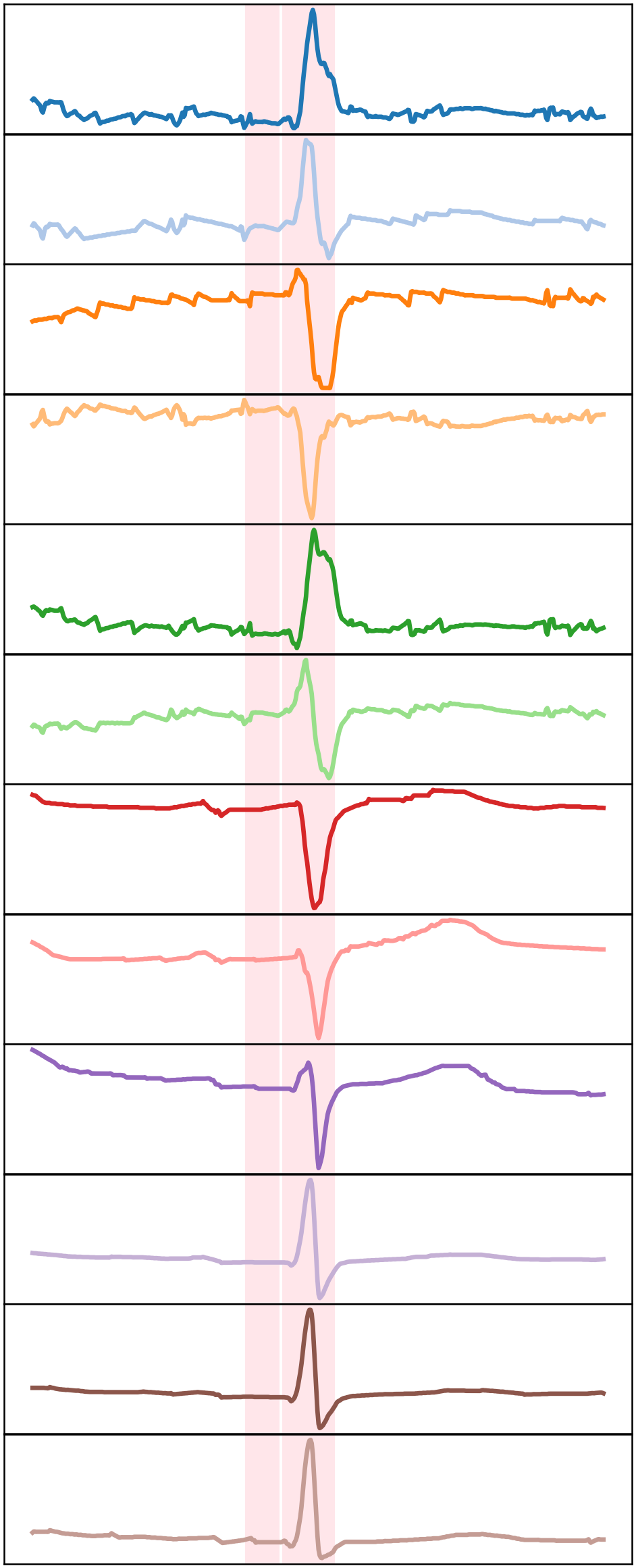}
    \caption{}
  \end{subfigure}
  \hfill
  \begin{subfigure}[t]{0.35\linewidth}
    \centering
    \includegraphics[width=0.95\linewidth]{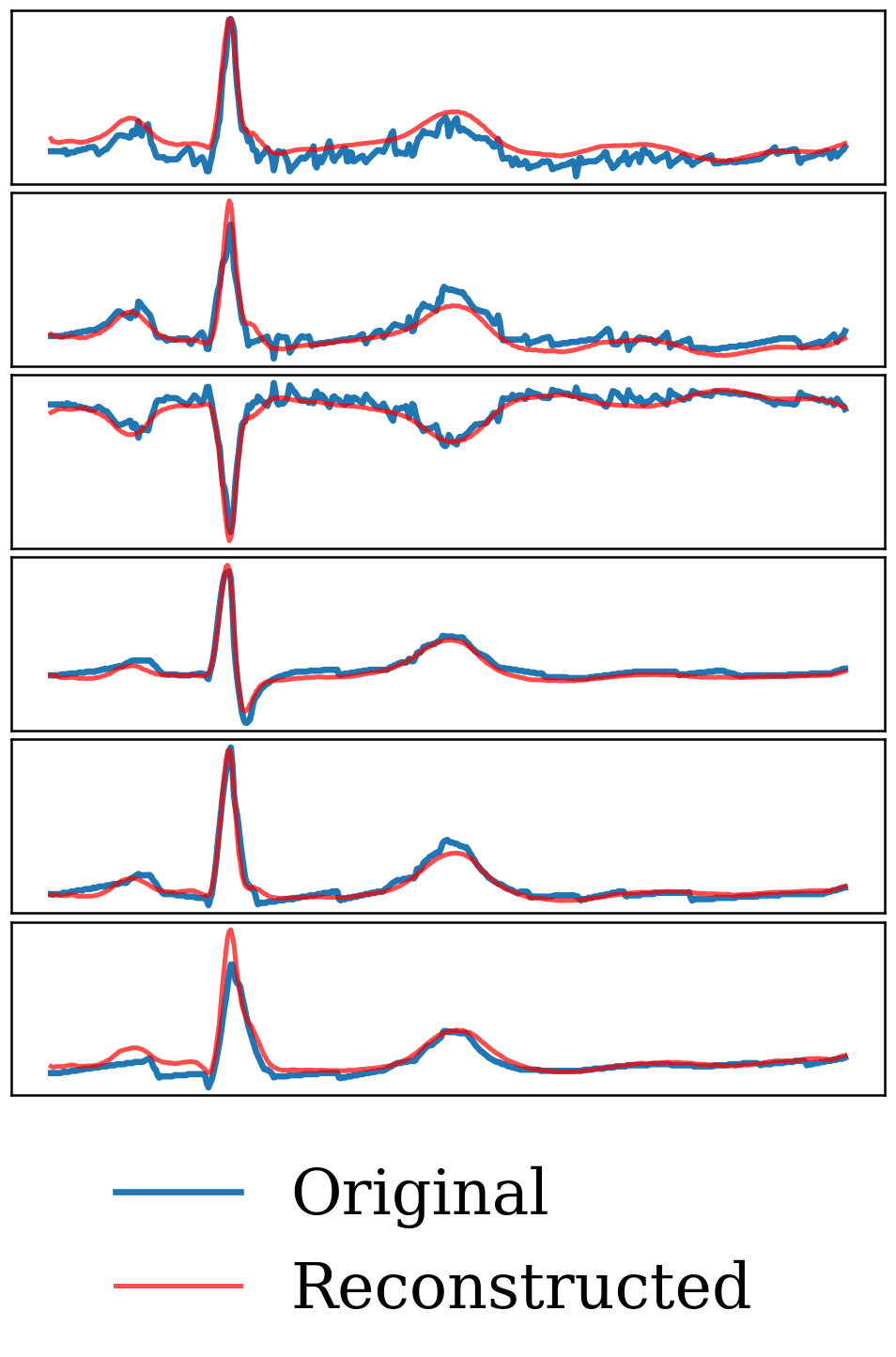}
    \caption{}
  \end{subfigure}

\caption{Visualization of latent alignment and signal reconstruction. (a-b) show attention maps for clinical queries, with shaded areas denoting high-impact segments. (c) shows reconstructed waveforms from symbolic tokens, demonstrating that discretization suppresses noise overfitting while preserving key information.}
  \label{fig:ecg_attn_vis}
\end{figure}

\paragraph{Evaluating Latent Alignment and Signal Reconstruction.}
To assess whether the model captures clinically meaningful regions, we perform token-masking attribution: sequentially masking each ECG token and measuring the resulting drop in EM ACC to estimate its importance, which is then mapped back onto the original ECG waveform to visualize model attention. Figure~\ref{fig:ecg_attn_vis}(a) shows that when asked about \textit{non-diagnostic T abnormalities}, the model emphasizes the T-wave segments. In contrast, for the \textit{myocardial infarction} query in Figure~\ref{fig:ecg_attn_vis}(b), attention shifts toward P and QRS complexes—consistent with diagnostic criteria. This context-dependent focus suggests that our alignment strategy successfully links ECG segments with query semantics. Additionally, Figure~\ref{fig:ecg_attn_vis}(c) demonstrates that discretization acts as an implicit regularizer, suppressing overfitting to noisy ECG segments while preserving clinically informative waveform morphologies through accurate reconstruction.

\paragraph{Cross-Dataset Generalization via Discretization.}
To examine the robustness of our learned representations, we evaluate zero-shot generalization by training on MIMIC-IV-ECG and directly testing on PTB-XL. As reported in Table~\ref{tab:ptbxl_zero_shot}, our method achieves the highest QA-Verify and QA-Query scores among all baselines, and the best average EM accuracy overall. Notably, although ST-MEM slightly outperforms our model on QA-Choose, its QA-Query performance remains weak. In contrast, our discretized token representation provides more balanced gains across all QA subtasks. These results indicate that the proposed symbolic ECG representations, implicitly aligned with the LLM’s textual embedding space, enhance domain transfer and resist overfitting to dataset-specific artifacts.

\begin{table}[t]
\centering
\setlength{\tabcolsep}{3pt}
{\fontsize{7}{8}\selectfont
\begin{tabular}{lcccc}
\toprule
\textbf{Methods} & \textbf{QA-Verify} & \textbf{QA-Choose} & \textbf{QA-Query} & \textbf{QA-Average} \\
\midrule
TIMELLM           & 66.81 & 26.62  & 11.62 & 35.02 \\
TEST              & 31.96 & 26.45  & 1.40  & 19.94 \\
MEIT              & 66.60 & 25.75  & 7.02  & 33.12 \\
LLMTIME           & 66.50 & 37.88  & 7.22  & 37.20 \\
ChatTime          & 67.66 & 39.95  & 6.59  & 38.07 \\
ST-MEM            & 67.61 & \textbf{41.08} & 7.94  & 38.88 \\
W2V-CMSC-RLM      & 61.82 & 36.46  & 4.85  & 34.38 \\
w/o DISC          & 64.93 & 37.32  & 5.06  & 35.77 \\
\textbf{Ours}     & \textbf{68.86} & 39.56 & \textbf{13.70} & \textbf{40.71} \\
\bottomrule
\end{tabular}
}
\caption{Zero-shot performance on PTB-XL using models trained on MIMIC-IV-ECG.}
\label{tab:ptbxl_zero_shot}
\end{table}

\section{Conclusion}

We present \textit{HeartLLM}, a unified framework that enables language-based clinical reasoning over ECG signals through symbolic tokenization and lightweight adaptation. By discretizing multi-lead ECG signals into a structured vocabulary, HeartLLM bridges the modality gap between continuous physiological data and discrete language representations. It avoids reliance on explicit ECG-text alignment or contrastive supervision, instead leveraging autoregressive pretraining and instruction tuning to support open-ended tasks such as question answering and report generation. Extensive experiments across multiple benchmarks demonstrate that HeartLLM achieves state-of-the-art performance and generalizes well to unseen ECG distributions. 
While effective, HeartLLM assumes offline processing and does not support real-time ECG analysis. Future work may extend our approach to streaming settings and incorporate medical knowledge to improve interpretability.

\bibliography{aaai2026}
\end{document}